\documentclass[letterpaper, 10 pt, conference]{ieeeconf}  

\usepackage[
  expansion=alltext,
  protrusion=alltext-nott, 
  final 
]{microtype}
\usepackage{amsmath}
\usepackage{amssymb}
\usepackage{dsfont}
\usepackage{siunitx}
\usepackage{graphicx, import}
\usepackage{cite}  
\usepackage{booktabs}
\usepackage{bm}
\usepackage{capt-of} 
\usepackage{placeins} 
\usepackage[nohyperlinks, nolist]{acronym}
\usepackage[hidelinks]{hyperref}
\usepackage{xcolor}
\usepackage{soul}
\usepackage{cancel}
\usepackage[normalem]{ulem}
\graphicspath{{figures}}
\newcommand{\rev}[1]{{\color{blue}#1}}
\newcommand{\del}[1]{%
  {\color{blue}%
  \ifmmode
    \cancel{#1}%
  \else
    \sout{#1}%
  \fi}%
}
\renewcommand{\rev}[1]{#1}
\renewcommand{\del}[1]{}

\newcommand*{\vect}[1]{\bm{#1}}

\IEEEoverridecommandlockouts

\title{\LARGE \bf
  Coverage Planning for Robotic Tooth Preparation in Densely Constrained Environments
}

\author{\del{Anonymous Authors}
\rev{Yunwen Li, Chen Chen, Xiangjie Yan, Chang Shu, Jianxia Hou, Shiji Song, and Xiang Li}
\thanks{\rev{Y. Li, C. Chen, X. Yan, S. Song, and X. Li are with the Department of Automation, Tsinghua University, China. Y. Li is also with the Department of Mechanical and Process Engineering, ETH Zürich, Switzerland. C. Chen and X. Yan are also with Tsingscribe Medical Ltd., China. C. Shu and J. Hou are with the Department of Priority Oral Care, Peking University School and Hospital of Stomatology. This work was supported 
in part by the Fundamental and Interdisciplinary Disciplines Breakthrough Plan
of the Ministry of Education of China under Grant JYB2025XDXM208, 
in part by the National Natural Science Foundation of China under Grant 62461160307, 
and in part by Beijing National Research Center for Information Science and Technology. 
Corresponding author: Xiang Li (xiangli@tsinghua.edu.cn)}}
}

\begin{document}

\maketitle
\thispagestyle{empty}
\pagestyle{empty}

\begin{abstract}
Tooth preparation refers to the controlled removal of tooth structure to create an optimal substrate for fixed restorations and is a core procedure in restorative dentistry. Automating this task is particularly challenging for robots because the dental bur must operate within a densely constrained intraoral workspace, where even sub-millimeter deviations can compromise outcomes or damage adjacent structures. This paper presents a novel robotic system for autonomous full-crown tooth preparation. The proposed framework includes: 1) an anatomy-aware toolpath planning algorithm that conforms precisely to a technician-designed preparation model while protecting adjacent teeth, and 2) a clearance-oriented end-effector yaw assignment strategy that allows intraoral access while reducing the risk of soft-tissue interference. Together, these features enable the robot to accurately mill the irregular tooth surface with an average geometric deviation of \qty{0.117}{mm} (RMSE), achieving both restoration quality and clinical safety. A series of simulations and phantom-head experiments validate the system's feasibility and effectiveness.

\end{abstract}


\section{Introduction}

\begin{figure}[!t]
    \centering
    \includegraphics[width=6cm]{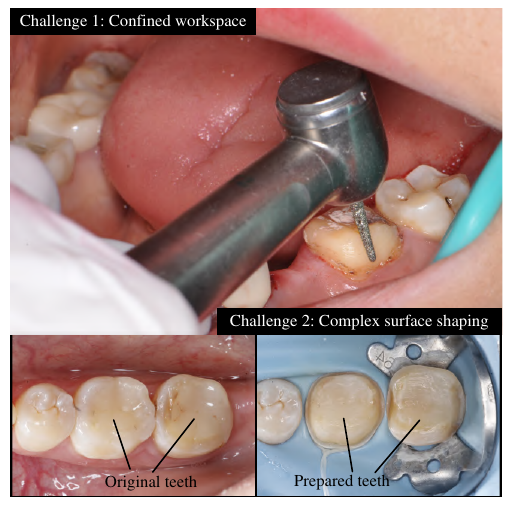}
    \caption{Task and key challenges in tooth preparation. The procedure is performed in a confined intraoral workspace and requires complex surface shaping, posing challenges to geometric accuracy and collision avoidance.}
    \label{fig:tooth_before_after}
\end{figure}

Tooth preparation refers to the controlled removal of tooth structure (see Fig.~\ref{fig:tooth_before_after}) to create an optimal environment for the placement of restoratives, ensuring both functional and aesthetic outcomes \cite{dilafruzToothPreparationCrucial2024}. It is a critical stage in restorative dentistry, serving as the first step for fabricating crowns, veneers, and other fixed restorations. The quality of tooth preparation significantly influences the longevity and performance of the final restoration. The procedure demands sub-millimeter accuracy in defining taper, finishing line, and surface smoothness \cite{goodacreToothPreparationsComplete2001}, while being performed in the confined and irregular environment of the oral cavity. Consequently, the procedure is typically performed manually by skilled dentists, with its quality largely dependent on the operator's expertise.

These challenges make tooth preparation not only technically demanding for clinicians but also a compelling target for robotic support. Dental robotics has emerged in recent years as a promising direction to improve precision, reproducibility, and ergonomics in dental surgical procedures \cite{liuRoboticsDentistryNarrative2023}. Among these, implant surgery robots \cite{wangAccuracyYakebotDental2024, boldingAccuracyHapticRobotic2022, yangAccuracyAutonomousRobotic2023} have reached the highest level of clinical maturity. Robotic assistance has also been investigated in areas such as tooth preparation \cite{ma3DPathPlanning2014, yuanAutomaticToothPreparation2016, yuanPreliminaryStudyTooth2020, sunLayeredPreparationMethod2024, sunOptimizationGrindingParameters2025a, sunTrajectoryOptimizationTooth2025, sunDigitalInteractiveDesign2023}, tooth cleaning \cite{sakaedaDevelopmentAutomaticTeeth2017}, oral and maxillofacial surgery \cite{chaoPreprogrammedRoboticOsteotomies2016}, and orthodontics \cite{liuReviewResearchProgress2024}, addressing different stages of the treatment workflow and exhibiting varying levels of clinical integration.

Despite recent progress in dental robotics, autonomous tooth preparation remains particularly challenging. The task requires high-precision material removal in a confined workspace, where sub-millimeter errors can affect restoration quality. In contrast to implant surgery with relatively simple drilling paths, tooth preparation involves complex three-dimensional surface shaping. In addition, limited access to posterior teeth and the presence of adjacent teeth and soft tissues impose strict constraints on tool motion. These requirements place high demands on trajectory planning and tool control for safe and accurate robotic execution.

\begin{figure*}
  \centering 
  \includegraphics[width=\linewidth]{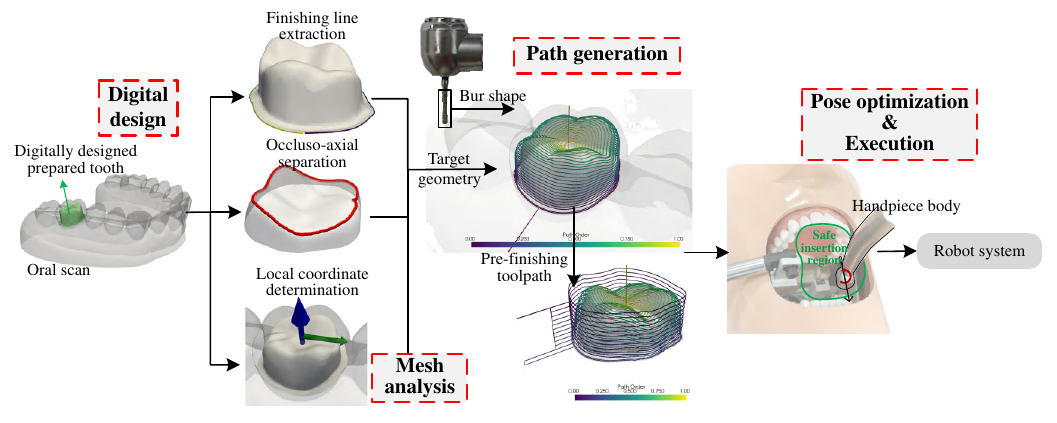}
  \caption{Overview of the proposed robotic tooth preparation framework. The system integrates (1) a technician-designed digital \del{crown }model as the preparation target, (2) anatomy-aware toolpath planning that respects bur geometry and adjacent-tooth safety, and (3) six-axis robotic execution with optimized handpiece insertion to ensure feasible intraoral access and precise material removal.}
  \label{fig:framework}
\end{figure*}

As a result of these challenges, existing robotic systems for tooth preparation remain constrained either by preparation type (e.g., the laser-based approach reported in \cite{yuanPreliminaryStudyTooth2020} was limited to shoulder-less crown preparation) or by operating on isolated teeth without considering adjacent teeth and soft-tissue interference in robotic-arm approaches \cite{sunTrajectoryOptimizationTooth2025}. Bridging this gap requires a system that combines clinically relevant tool usage, accurate and safe trajectory planning, and feasible intraoral access. Therefore, we present a robotic system capable of performing autonomous full-crown tooth preparation with anatomy-aware planning and clinically feasible execution. The main contributions of this work are as follows:
\begin{itemize}
    \item We propose a novel robotic framework for autonomous full-crown tooth preparation that prepares the tooth to match a technician-designed geometry. This strategy naturally reflects the digital restorative dentistry workflow, where restorations are planned first and subsequently manufactured.
    \item We develop an anatomy-constrained coverage planning algorithm for toolpath generation that accounts for the geometry of the target preparation, the shape of a conventional dental bur, and the protection of adjacent teeth, thereby improving both accuracy and safety.
    \item We extend the five-DOF preparation path to full six-DOF motion by \del{defining a heuristic handpiece insertion direction that reduces the risk of interference with surrounding soft tissues, such as the buccal mucosa, in the confined oral cavity}\rev{assigning the residual yaw about the bur axis so that the handpiece body points outward from the oral cavity, thereby reducing the risk of interference with surrounding soft tissues}.
\end{itemize}

The developed robot is validated through both simulations and experiments on dental phantom models, demonstrating its feasibility and effectiveness.


\section{Related Work}
Robotic tooth preparation has been explored through two main approaches: laser-based ablation and handpiece-based robotic arms. Ma et al. \cite{ma3DPathPlanning2014} first proposed a 3D path-planning method using an ultra-short pulse laser. Building on this, Yuan et al. \cite{yuanAutomaticToothPreparation2016,yuanPreliminaryStudyTooth2020} demonstrated a robot-controlled ultra-short pulse laser system capable of automatic crown preparation, confirming its accuracy and feasibility. However, this system is limited to shoulder-less full-crown designs \cite{yuanPreliminaryStudyTooth2020} and faces challenges such as temperature control and surface discoloration during ablation \cite{sunLayeredPreparationMethod2024}.

More recently, Sun et al. have developed a robotic tooth preparation system based on a conventional dental handpiece mounted on a robotic arm. They first introduced a digital interactive design method for veneer preparation \cite{sunDigitalInteractiveDesign2023}. Subsequently, they proposed a layered full-crown preparation strategy \cite{sunLayeredPreparationMethod2024}, in which tooth material is removed incrementally until the desired depth is achieved. Their later studies focused on process optimization: grinding parameters were analyzed using fracture mechanics theory and a multi-grit thermal-mechanical coupling finite element model \cite{sunOptimizationGrindingParameters2025a}; trajectory generation was improved through the P-MRSD method, which predicts material residue and stiffness deformation based on parameters such as bur pose and preparation trajectory, resulting in higher preparation accuracy \cite{sunTrajectoryOptimizationTooth2025}. Furthermore, extrusion force during preparation was optimized by considering factors such as material removal rate, tool-tooth contact fields, and system stiffness. However, these studies were primarily evaluated on isolated teeth and did not address collision risks with adjacent teeth or soft tissues during intraoral execution \cite{sunTrajectoryOptimizationTooth2025}. Further addressing these issues is essential for achieving clinically viable robotic tooth preparation.


In parallel, tooth preparation can be viewed as a freeform surface milling problem performed in a highly confined workspace. Toolpath generation has been extensively studied in the context of freeform surface machining, with a comprehensive survey provided in \cite{lasemiRecentDevelopmentCNC2010}. Traditional CNC machining is usually performed in an open environment, where the tool can retract and reorient freely to avoid collision, prevent gouging, and ensure surface quality. In contrast, tooth preparation occurs in an irregular oral cavity bounded by neighboring teeth, gingiva, and buccal mucosa—anatomical structures that are difficult to model accurately in advance. \del{For instance, although using only the tip of an end mill can achieve precise point-based tool--object contact for accurate shaping, this strategy is infeasible and inefficient intraorally; clinicians instead rely on the lateral surface of a dental bur to remove material.}\rev{For instance, using only the tip of an end mill with the tool axis continuously aligned to the local surface normal enables precise point-based tool--object contact for accurate shaping. However, this strategy is infeasible intraorally. Instead, clinical tooth preparation primarily relies on the lateral surface of a dental bur, with the bur orientation remaining close to the tooth's insertion direction.} There is virtually no prior work on adapting freeform surface machining to such narrow conditions. This gap motivates the development of anatomy-aware path planning strategies specifically tailored to the oral cavity.


\section{System Overview}
\begin{figure}[ht]
  \centering
  \includegraphics{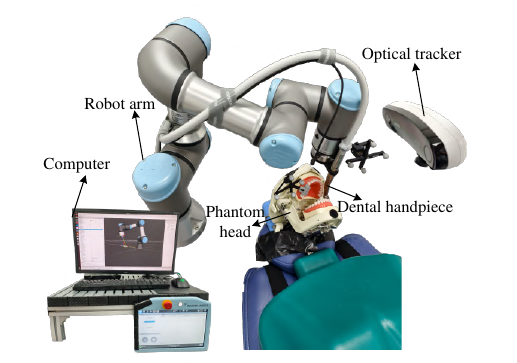}
  \caption{The prototype of the tooth preparation robot.}
  \label{fig:expsetup}
\end{figure}
Recent advances in digital dentistry—particularly intraoral scanning, CAD/CAM design, and chairside fabrication—enable restorations to be digitally planned before material removal \cite{rekowDigitalDentistryNew2020}. Leveraging this workflow, our system begins with a technician-designed target tooth preparation model and an intraoral scan of the dentition. An overview of the proposed framework is shown in Fig.~\ref{fig:framework}. Specifically: 1) \textbf{Digital Design} acquires the intraoral scan using a TRIOS 3 intraoral scanner (3Shape, Copenhagen, Denmark) and the technician-designed preparation model; 2) \textbf{Mesh Analysis} is performed to extract the finishing line and establish a local tooth frame, as well as to separate occlusal and axial surfaces for (pre)molar teeth, providing the geometric foundation for subsequent path planning; 3) \textbf{Path Generation} creates precise toolpaths from the target tooth geometry while respecting anatomical constraints; 4) \textbf{Handpiece Pose Optimization} refines the orientation of the handpiece to reduce the risk of collision with surrounding soft tissues and maintains anatomically feasible execution.

This paper develops a new tooth preparation robot, as shown in Fig.~\ref{fig:expsetup}, mainly consisting of: 1) an optical tracking system, which registers the transformation between the robot arm and the target tooth, 2) a 6-DOF robot manipulator (Universal Robots UR3e) with a traditional high-speed dental handpiece mounted as the end-effector, \del{3) an ATI mini40 force/torque (FT) sensor mounted between the arm flange and the dental handpiece,} \del{4}\rev{3}) a dental chair, and \del{5}\rev{4}) a computer for processing the data and controlling the manipulator.

\section{Methods}
Effective toolpath planning for robotic tooth preparation must satisfy three critical constraints:

\begin{itemize}
    \item \textbf{Precision} - ensure accurate material removal so the prepared tooth faithfully matches the technician's design.
    \item \textbf{Adjacent-Teeth Protection} - avoid damaging neighboring teeth; in our method, this is achieved geometrically by referencing the finishing line as a safety margin.
    \item \textbf{Anatomical Safety} - maintain feasible handpiece orientations to prevent collisions with surrounding soft tissues (e.g., buccal mucosa) within the restricted oral cavity.
\end{itemize}

The following subsections describe the analysis and planning steps used to generate toolpaths that meet them.

\subsection{Mesh Analysis}
\label{sec:mesh_analysis}
\begin{figure}[t]
      \centering
      \includegraphics{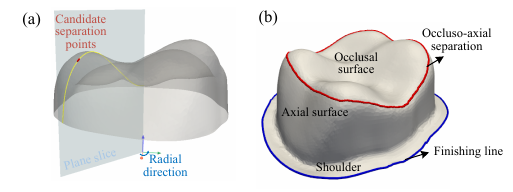}
      \caption{Occluso-axial separation. (a) Candidate points for each radial direction in the dynamic programming algorithm. (b) Finishing line and occluso-axial separation.}
      \label{fig:mesh_analysis}
    \end{figure}

	After obtaining the digital models of both the original tooth and the target prepared tooth, we generate an executable preparation path by first analyzing the three-dimensional morphology of the mesh. The digital models are represented as triangular meshes, where the surface consists of a large number of adjacent triangular facets, and the facet normal vectors describe the local surface orientation.
    \paragraph{Finishing line extraction} The finishing line is the margin on the prepared tooth that marks the boundary between the prepared surface and the unprepared tooth structure, i.e., the intersection curve between prepared and original tooth. It is extracted from the designed prepared tooth by identifying edges that are not shared by two faces (i.e., boundary edges, see Fig.~\ref{fig:mesh_analysis}(b)). It serves a dual role in our planning: \\
1) as the geometric seed for generating the first shoulder-forming layer of the toolpath, and \\ 
2) as a hard \textit{safety boundary} to protect adjacent teeth.
    \begin{figure}[t]
      \centering
      \includegraphics{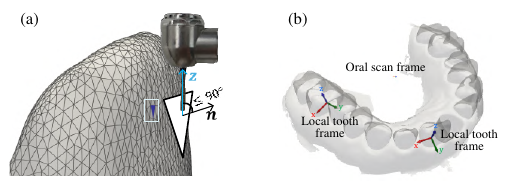}
      \caption{Local tooth frames for toolpath generation. (a) Criteria for $z$-axis optimization. (b) Sample local frames in oral scan frame.}
      \label{fig:local_frame}
    \end{figure}

    \paragraph{Occlusal-axial separation}
    For molar and premolar teeth, the occlusal and axial surfaces are separated by extracting a closed loop through dynamic programming. Candidate points above the tooth shoulder are first sampled along radial slicing planes and ranked using a slope criterion to detect the transition from occlusal to axial regions (see Fig.~\ref{fig:mesh_analysis}(a)). A shortest-path optimization is then applied to connect one candidate from each slicing plane into a smooth, closed loop. This loop consistently follows the upward-facing facets whose normals form an angle of roughly $45^{\circ}$ upward, thereby serving as the occlusal-axial boundary.

    \paragraph{Local frame determination}
    For simplicity, the bur orientation is kept constant with respect to the target tooth throughout preparation. Specifically, an optimal tool axis direction is determined in advance and used to define the $z$-axis of a local tooth frame. All toolpaths are subsequently generated in this frame, as illustrated in Fig.~\ref{fig:local_frame}(b).
    
    As shown in Fig.~\ref{fig:local_frame}(a), to avoid damaging a target facet during cutting, the bur direction $\vect{z}$ must form an angle no greater than \qty{90}{\degree} with the facet normal $\vect{n}$. Therefore, the optimal tool axis $\vect{z}$ is obtained by solving the following optimization problem:
	\begin{equation}
	\vect{z} = \arg\max_{\vect{z} \in S^2} \sum_{i=1}^N \mathds{1}(\vect{z} \cdot \vect{n_i}\ge 0),
  \label{eq:tool_axis_binary}
	\end{equation} 
	where $N$ is the number of faces in the target prepared tooth mesh, and $\mathds{1}(\cdot)$ is the indicator function, which equals 1 if the condition holds and 0 otherwise. Geometrically, this formulation searches for a direction on the unit sphere that forms a non-obtuse angle with as many facet normals as possible, thereby determining the optimal tool axis with respect to the tooth.
	
	Since the indicator function is non-differentiable and unsuitable for gradient-based optimization, we adopt a smooth approximation using the Sigmoid function. The differentiable objective function is then formulated as:
    \begin{equation}
    \vect{z} = \arg\max_{\vect{z} \in S^2} \sum_{i=1}^N \frac{1}{1 + e^{-k(\vect{z} \cdot \vect{n_i})}}\del{,}\rev{.}
    \end{equation}
    \del{where $k$ is a tunable smoothing factor controlling the steepness of the Sigmoid function.}\rev{The smoothing factor $k$ is set to $100$. For this value, the Sigmoid decreases from $0.95$ to $0.05$ as the angle between $\vect{z}$ and $\vect{n_i}$ varies from approximately \qty{88.3}{\degree} to \qty{91.7}{\degree}, closely approximating the binary objective in~(\ref{eq:tool_axis_binary}). Since clinically designed target prepared teeth have a tapered geometry around the tooth insertion direction, maximizing (\ref{eq:tool_axis_binary}) naturally yields a tool axis close to the insertion direction.} Once the tool axis direction $\vect{z}$ is determined, a lateral direction $\vect{y}$ orthogonal to $\vect{z}$ is defined to complete the local coordinate system for bur motion. This lateral direction is manually selected using a graphical user interface (GUI). It is always selected towards the outside of the mouth, and will later be used to optimize the pose of the dental handpiece.

    For premolars and molars, the origin is defined as the center of the finishing line projected downward along the $z$-axis to the lowest $z$-coordinate of the finishing line. For incisors and canines, the origin is instead defined as the highest point on the mesh projected downward to the lowest $z$-coordinate of the finishing line.
	
\subsection{Precise Path Generation}
\label{sec:path_generation}
    \begin{figure}[ht]
      \centering
      \includegraphics{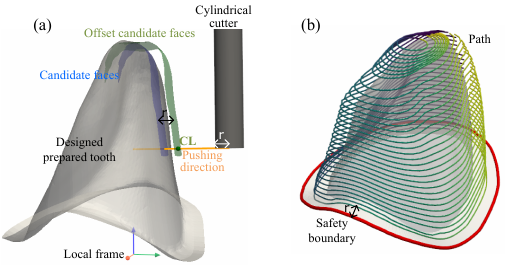}
      \caption{Toolpath generation for incisors, canines, and the axial surfaces of (pre)molar teeth. (a) Geometric model for calculating facet contact in each pushing direction. The candidate facets are offset by $r$ in their normal direction projected onto the $XY$ plane. (b) Cutter location result for all layers and pushing directions. The finishing line is inwardly offset by the cutter radius $r$ and added as \rev{the} first layer, providing both the initial shoulder shape and a safety boundary for adjacent-teeth protection.}
      \label{fig:geo_calc}
    \end{figure}

   Our path generation focuses on a purely \textit{geometric model} to guarantee high \textbf{precision}. During the entire preparation process, the bur is kept parallel to the previously optimized $z$-axis.
   
   \subsubsection{Incisors, Canines, and Axial Surfaces}
   For incisors, canine teeth, and the axial surfaces of premolars and molars, a bottom-up layer-by-layer toolpath strategy is adopted. Starting from the finishing line, closed loops are generated layer by layer until the complete target geometry is reached.
	
    Instead of using conventional iso-height slicing, the toolpath is determined by sampling along radial directions. For each angular direction, the height difference between the finishing line and the uppermost geometry is evaluated, resulting in a path whose vertical position varies with the angular coordinate. This non-uniform sampling ensures that the trajectory conforms to the actual morphology of the finishing line rather than to an idealized planar slice.
	
    For these tooth surfaces, the toolpath is generated by pushing the cutter into the model along discrete angular directions in the $XY$ plane. The computation follows a purely geometric approach: for each pushing direction, we determine the cutter contact (CC) point on the prepared tooth surface and its corresponding cutter location (CL) point on the tool axis (i.e., the tip center for a flat-end cutter). In this study, we employ a cylindrical flat-end cutter with a diameter of \qty{1.0}{mm} (SF-10) to create a flat shoulder. This geometry simplifies the formulation of the contact equations presented below. Nevertheless, the proposed framework is not limited to cylindrical cutters and can be extended to other tool geometries, such as ball-end or conical burs, by modifying the corresponding CC-CL geometric relations.
    
    Along each pushing path, only a subset of mesh triangles may intersect with the cutter. These candidate facets (including edges and vertices) are examined in detail to calculate the precise CL position:
	
    \paragraph{Facet contact}  
    For a directional path from $\vect{p_1}$ to $\vect{p_2}$, let a CL point on the path be 
	\begin{equation}
	\vect{p}(\lambda) = \vect{p_1} + \lambda(\vect{p_2} - \vect{p_1}), \qquad \lambda \in [0,1].
	\end{equation}
	For a triangular facet with vertices $\vect{v_0}$, $\vect{v_1}$, $\vect{v_2}$, let a CC point on the facet be 
    
    \begin{equation}
    \rev{
    \begin{gathered}
    \vect{q}(a,b)
    = \vect{v_0}
    + a(\vect{v_1}-\vect{v_0})
    + b(\vect{v_2}-\vect{v_0}), \\
    a\geq 0,\qquad b\geq 0,\qquad a+b\leq 1.
    \end{gathered}
    }
    \end{equation}
    
	For a cylindrical cutter, the facet contact $a$, $b$, $\lambda$ can be solved using
	\begin{equation}
	 \vect{q}(a, b) + r\vect{n}_{xy} = \vect{p}(\lambda),
	\end{equation}
	where $\vect{n}_{xy}$ is \del{the facet normal projected}\rev{the normalized projection of the facet normal} onto the $XY$ plane, and $r$ is the cutter radius. Fig.~\ref{fig:geo_calc}(a) provides an example.

    \paragraph{Edge contact}  
    For an edge from $\vect{v_0}$ to $\vect{v_1}$, let a point on the edge be 
    \begin{equation}
    \vect{g}(t) = \vect{v_0} + t(\vect{v_1} - \vect{v_0}), \qquad t \in [0,1].
    \end{equation}
    Denote its projection onto the $XY$ plane and its being lifted to the path height $z$ as $\Pi_{xy}(\vect{g}(t),z)$. The cutter contacts the edge when the horizontal distance between the cutter axis and this lifted point equals the cutter radius $r$:
    \begin{equation}
        \big\| \Pi_{xy}(\vect{g}(t), z) - 
        \vect{p}(\lambda) \big\|_2 
        = r.
    \end{equation}
    Or, for mathematical simplicity, edge contact can equivalently be computed by offsetting the projected edge in its local normal direction by $r$ and intersecting it with the pushing path.
    
    \paragraph{Vertex contact}  
    The contact between the cylindrical cutter and a triangular vertex $\vect{v_0}$ can be expressed as
    \begin{equation}
        \big\| \Pi_{xy}(\vect{v_0},z) - \vect{p}(\lambda) \big\|_2 
        = r.
    \end{equation}

    \del{By combining these three cases, the earliest CC point and the corresponding CL along each pushing direction can be computed.}\rev{By combining these three cases, the earliest contact is selected as the valid solution with the minimum $\lambda$ along the pushing direction, and the corresponding $\vect{p}(\lambda)$ is taken as the cutter location.} This ensures accurate toolpath generation and precise conformity to the target mesh surface. The toolpath can then be generated for all radial directions for all layers (see Fig.~\ref{fig:geo_calc}(b)).

    \subsubsection{Occlusal Surfaces}
    \begin{figure}[ht]
      \centering
      \includegraphics{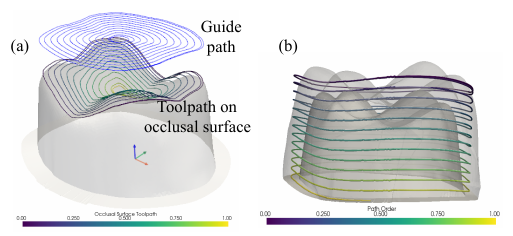}
      \caption{(a) Toolpath for the occlusal surfaces of (pre)molar teeth. (b) Pre-finishing toolpath for (pre)molar teeth.}
      \label{fig:guide_prefinishing}
    \end{figure}
    For the occlusal geometry of premolar and molar teeth, a similar geometric approach is used. However, instead of pushing cutters in the $XY$ plane, we drop the cutter from above (downward along the $z$-axis) until it touches the designed prepared tooth. A spiral guide path is generated from outside to inside (see Fig.~\ref{fig:guide_prefinishing}(a)) from which the cutter drops. For a smoother transition between axial and occlusal surfaces, we further interpolate several layers between the uppermost axial surface toolpath and the \del{outside-most}\rev{outermost} occlusal surface toolpath.
    
    \rev{The geometric contact calculations described in this subsection were implemented using a modified version of OpenCAMLib~\cite{opencamlib}.}

\subsection{Adjacent-Teeth Protection}
\label{sec:adj_protect}
Although the primary objective of path generation is geometric precision, the cutter motion must also respect the anatomical context of the oral cavity---specifically, the need to \textbf{protect adjacent teeth}. During preparation, even small overshoots of the toolpath beyond the target tooth margin can result in unintended contact with neighboring teeth.

To satisfy this anatomical constraint, we use the finishing line extracted previously as a safety boundary. Starting from the bottom layer, the finishing-line boundary is first offset inward by the cutter radius $r$, as indicated in Fig.~\ref{fig:geo_calc}(b). Each higher layer is then checked so that its toolpath remains entirely within the footprint of the layer beneath it. \rev{In regions with insufficient interproximal clearance, the cutter is allowed to penetrate the target preparation so that the cutter surface remains within the finishing-line boundary, sacrificing local geometric conformity to protect adjacent teeth. Since the tool axis optimized using~(\ref{eq:tool_axis_binary}) remains close to the tooth insertion direction, this inward offset substantially reduces the risk of interference with adjacent teeth.} Lastly, different layers are connected bottom up using angular spiral interpolation.

\begin{figure}[ht]
    \centering
    \includegraphics[]{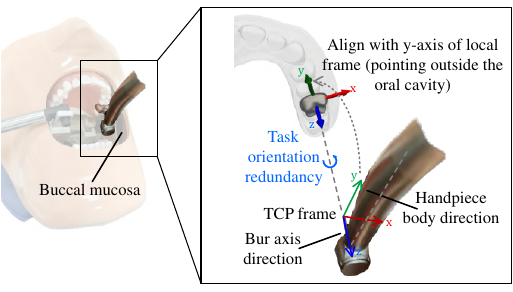}
    \caption{Definition of the handpiece orientation for implicit obstacle avoidance.  
By aligning the handpiece $y$-axis (perpendicular to the bur axis and maximally aligned with the physical longitudinal direction of the handpiece body)  
with the outward-pointing $y$-axis of the tooth frame, the five-DOF toolpath is extended to a six-DOF trajectory that maintains a safe orientation relative to the buccal mucosa.}
    \label{fig:pose_opt}
\end{figure}

\subsection{Pre-finishing Toolpath}
    Because the toolpath is generated directly from the target prepared tooth geometry, certain regions may require removing a relatively large amount of material from the original tooth. This can increase the risk of the bur binding. To mitigate this, we introduce an additional pre-finishing toolpath prior to the planned accurate toolpath (see Fig.~\ref{fig:guide_prefinishing}(b)). In this pre-finishing stage, the offset finishing line is lifted to multiple heights to form successive layers, which are arranged from top to bottom and interpolated spirally to gradually remove material. Each layer is lifted from the finishing line shrunk by $r$, protecting adjacent teeth. \del{Similar strategies can be adopted for incisors and canines.}\rev{For incisors and canines, to reduce preparation time, pre-finishing is limited to top-down passes on the proximal surfaces, with a few additional outside-in passes on the labial and lingual surfaces.}

\subsection{Handpiece Pose Optimization}
In addition to maintaining a safe distance between the bur and adjacent teeth---as discussed in Sec.~\ref{sec:adj_protect}---the manipulated handpiece must also avoid unintended contact with other intraoral structures, particularly the buccal mucosa (see Fig.~\ref{fig:pose_opt}). Because comprehensive geometric data of the entire intraoral environment are unavailable, a fully \textit{explicit} obstacle-avoidance strategy based on complete environment modeling cannot be implemented. Instead, \del{we enforce an additional dimension during trajectory planning,}\rev{we exploit the remaining yaw redundancy about the bur axis to orient the handpiece body toward the outside of the oral cavity,} enabling an \textit{implicit} avoidance strategy that reduces the risk of collisions with surrounding soft tissues.

During tooth preparation, the bur follows a precomputed toolpath whose position and orientation determine five of the handpiece's six degrees of freedom: the bur tip's 3D position and its orientation (pitch and roll) are fully defined by the preparation task. However, there is one remaining freedom \del{around}\rev{about} the bur's own $z$-axis (i.e., its local yaw). \rev{This redundancy does not alter the bur tip position or tool-axis direction; it only determines the handpiece body orientation around the bur axis.} Equivalently, when constructing the TCP (Tool Center Point) frame of the end-effector (handpiece plus bur) in the robot's flange frame, this residual yaw corresponds to how we choose the $x$-$y$ orientation around the tool axis $z$.
To represent the handpiece body consistently, we define its $y$-axis to be:  
\begin{enumerate}
    \item perpendicular to the bur axis ($z$-axis), and  
    \item among all such directions, it is chosen to be maximally aligned with the physical longitudinal direction of the handpiece body, which in our setup corresponds to the direction toward the robot flange.
\end{enumerate}

Under this convention, the handpiece's $y$-axis provides a consistent indication of its body pose. The tooth-frame's $y$-axis is defined to point outward the mouth (Sec.~\ref{sec:mesh_analysis}). We therefore extend the \textit{five-DOF} toolpath to a full \textit{six-DOF} trajectory by aligning the handpiece $y$-axis with the tooth-frame $y$-axis (see Fig.~\ref{fig:pose_opt}). In this way, the handpiece body strictly follows the prescribed orientation while executing the planned toolpath, reducing the likelihood of contact with surrounding structures.

\section{Results}
The proposed toolpath planning algorithm was validated in simulation on an incisor, a canine, and a molar tooth. A phantom-head experiment was subsequently performed on a molar tooth using a \qty{1}{mm}-diameter flat-end cylindrical bur (SF-10).
\subsection{Planning Results}
\begin{figure}[t]
  \centering
  \includegraphics{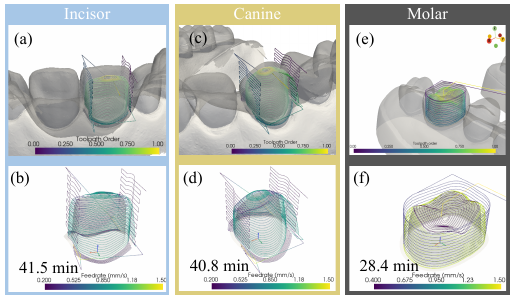}
  \caption{Complete toolpaths generated by the proposed planning algorithm for (a, b) incisor, (c, d) canine, and (e, f) molar preparation. The upper row shows the global toolpath distribution, while the lower row shows the feedrate profiles and the corresponding preparation time. The relatively long preparation time results from the conservative, safety-oriented feedrate selection.}
  \label{fig:planning_result}
\end{figure}
\begin{figure}[t]
  \centering
  \includegraphics{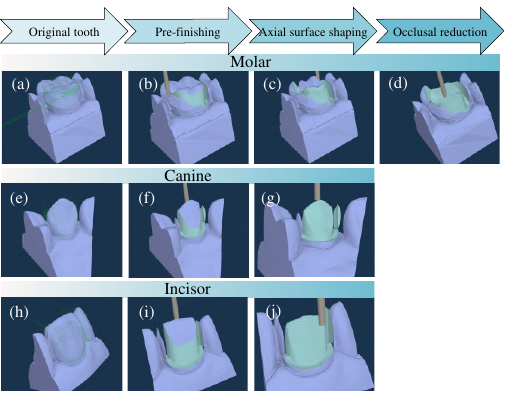}
  \caption{Simulated tooth preparation process on a molar (a--d), a canine (e--g), and an incisor (h--j).}
  \label{fig:sim}
\end{figure}
Fig.~\ref{fig:planning_result} demonstrates the complete toolpaths for preparing an incisor, a canine, and a molar tooth. The feedrates were selected empirically and conservatively. For the molar case in Fig.~\ref{fig:planning_result}, a staged feedrate profile was adopted: \qty{1.5}{\mm\per\s} for entry and exit, \qty{0.6}{\mm\per\s} for pre-finishing, \qty{1.4}{\mm\per\s} for axial shaping, and \qty{0.4}{\mm\per\s} for occlusal reduction, resulting in a preparation time of \qty{28.4}{\minute}. \rev{The relatively high feedrate for axial shaping is possible because only a small amount of residual material is removed after the pre-finishing stage.} Similar conservative feedrate strategies were applied to the incisor and canine cases, leading to preparation times of approximately \qty{41}{\minute}.

The preparation time is primarily determined by the selected number of layers and the feedrate profile along the toolpath. For safety considerations, the feedrate should be constrained by the amount of material removed in each layer. In future work, the overall preparation time can be optimized by jointly considering layer configuration, material removal rate, and cutting-force constraints.

\subsection{Simulation}
\begin{figure}[t]
  \centering
  \includegraphics{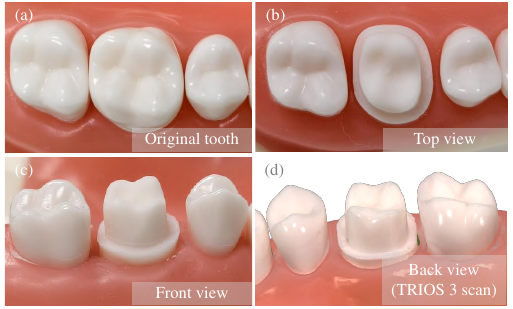}
  \caption{Original and prepared tooth on a phantom head.}
  \label{fig:exp}
\end{figure}
To validate the toolpath, we animated it in FreeCAD. For faster simulation, the original intraoral scan was cropped and simplified. As shown in Fig.~\ref{fig:sim}, the toolpath successfully prepared the target tooth to the intended shape. The finishing line was cleanly cut; although some residual material remained above and outside the finishing line, it would naturally detach under clinical conditions. Adjacent teeth were well protected using our proposed toolpath safety boundary.

\subsection{Experimental Validation}
\begin{figure}[t]
  \centering
  \includegraphics{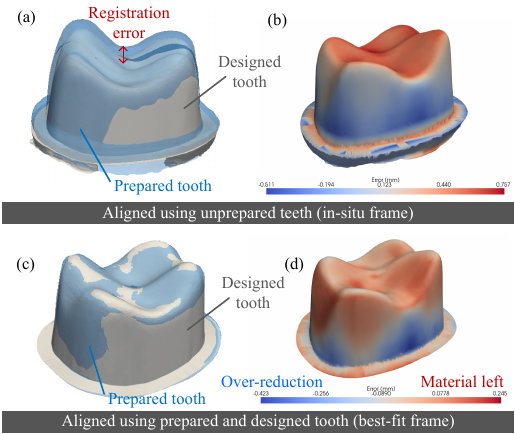}
  \caption{Surface deviation of the prepared tooth relative to the designed target. (a,b) Alignment based on other unprepared teeth. (c,d) Best-fit rigid alignment between the prepared and designed tooth.}
  \label{fig:result_analysis}
\end{figure}

\begin{table}[t]
\caption{Surface deviation metrics}
\label{tab:error_metrics}
\begin{center}
\begin{tabular}{ccc}
\hline
Metric & In-situ frame (\unit{\mm}) & Best-fit frame (\unit{\mm}) \\
\hline
Minimum signed error  & -0.511 & -0.423 \\
Maximum signed error & 0.757 & 0.245 \\
Root mean square error & 0.321 & 0.117 \\
90\% absolute deviation & 0.561 & 0.210 \\
\hline
\end{tabular}
\end{center}
\end{table}

Experiments were carried out on a molar tooth on a dental phantom, and the resulting prepared tooth is shown in Fig.~\ref{fig:exp}. \rev{Before execution, the target tooth was registered to the robot base using the optical tracking system. A CAD-designed rigid marker attached to the dental arch was tracked by the optical tracker. Combined with the calibrated transformation between the optical tracker and the robot base, this yielded the transformation from the preoperative oral scan to the robot base for trajectory execution.} After preparation, the dental arch was rescanned using the TRIOS 3 intraoral scanner and rigidly aligned with the preoperative oral scan. Two alignment references were used: an in-situ frame obtained by registering the post-preparation scan to the preoperative scan using unprepared teeth, and a best-fit frame obtained by rigidly aligning the prepared tooth to the designed target tooth.

To evaluate preparation accuracy, signed distances were computed by sampling points on the designed tooth mesh and measuring their closest-point deviation relative to the prepared tooth mesh. The sign of the error was determined by the surface normal of the designed mesh: positive values indicate regions where material remained unremoved, whereas negative values correspond to over-reduction.

Fig.~\ref{fig:result_analysis} shows the results under these two alignment strategies. Panels (a,b) show the mesh overlay and signed error map in the in-situ frame, reflecting error from both toolpath and robot-tooth registration. Panels (c,d) present the results after best-fit alignment, which emphasizes geometric discrepancy. The quantitative error metrics in both frames are summarized in Table~\ref{tab:error_metrics}.

In the in-situ frame, the root mean square error (RMSE) of \qty{0.321}{\mm} reflects accumulated toolpath, registration, and execution errors. After best-fit alignment, the RMSE decreases to \qty{0.117}{\mm}, and 90\% of surface points lie within \qty{0.210}{\mm} of the designed tooth, indicating that the intrinsic geometric deviation is clinically acceptable. Spatially, regions with near-zero values indicate accurate preparation, while larger deviations reflect local discrepancies introduced by the proposed safety boundary (which constrains the cutter to penetrate slightly into the target mesh), residual registration error, execution inaccuracy, and scanning noise.

Because the toolpath is generated in a layer-wise manner, a staircase-like surface texture is observable in low-resolution simulation (see Fig.~\ref{fig:sim}(c,g,j)) and can also be visually identified on the prepared tooth (see Fig.~\ref{fig:exp}(c)). \del{This discretization artifact is not prominent in the intraoral scan, and therefore does not affect the deviation metrics. The staircase effect can be mitigated by increasing the number of layers or by using a ball-end bur. Clinically, such surface roughness can be removed during final polishing and does not compromise the overall preparation quality.}\rev{This texture is not clearly resolved in the intraoral scan, likely because of scanner resolution and surface-reconstruction smoothing. Consequently, the reported mesh-based deviation metrics may not fully capture this local surface texture. The staircase effect can be reduced by increasing the number of layers or using a ball-end bur and can be further smoothed during the final finishing procedure.}

\begin{figure}[t]
  \centering
  \includegraphics{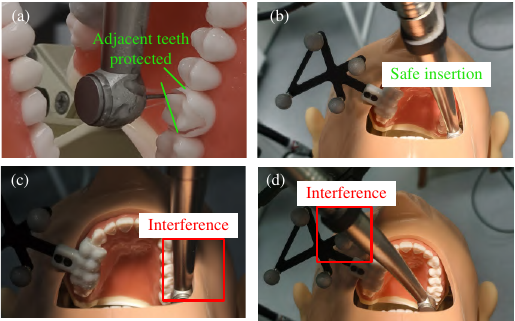}
  \caption{Anatomy-aware toolpath and clearance-oriented handpiece orientation. (a) Adjacent teeth were protected by the enforced toolpath boundary. (b) The proposed yaw selection strategy enables safe insertion. (c,d) Without the yaw assignment, the handpiece may interfere with the soft tissues and optical markers.}
  \label{fig:execution_safe}
\end{figure}
Fig.~\ref{fig:execution_safe} demonstrates the effect of the proposed toolpath boundary and handpiece yaw assignment strategy on execution safety. Adjacent teeth were protected by the safety boundary, and no interference with anatomical tissues or optical markers was observed when the proposed yaw selection strategy was applied.

\section{Conclusions}
We have presented an autonomous robotic system for full-crown tooth preparation that integrates anatomy-aware planning and clinically feasible execution. By generating toolpaths that conform to technician-designed preparation models and by defining a feasible handpiece orientation, the framework accommodates the confined intraoral workspace and the complex surface shaping required for restorative dentistry.
Phantom-head experiments and evaluations demonstrate that the proposed approach achieves sub-millimeter accuracy while maintaining safe distances from adjacent structures, indicating its potential for clinical translation. Future work will focus on improving the robustness and efficiency of robotic tooth preparation. This includes the development of more accurate registration methods and the optimization of feedrate scheduling, both offline during planning and online during execution, to achieve faster and more reliable operation.








\FloatBarrier
\bibliographystyle{IEEEtran}
\bibliography{IROS2026}

\end{document}